\documentclass[journal,twoside,web]{ieeecolor}
\usepackage{xcolor}
\usepackage{graphicx}
\usepackage{subcaption}
\usepackage{caption}
\definecolor{cerulean}{rgb}{0.0,0.48,0.65}
\definecolor{green}{rgb}{0.01, 0.75, 0.24}
\definecolor{Black}{RGB}{0.0, 0.0, 0.0}

\newcommand{\blue}[1]{\textcolor{blue}{#1}}

\def\s{\s}

\usepackage{generic}
\usepackage{cite}
\usepackage{amsmath,amssymb,amsfonts}
\usepackage{algorithmic}
\usepackage{graphicx}
\usepackage{booktabs}
\usepackage{multirow}
\usepackage{tabularx, colortbl} 
\newcommand{\shadow}[1]{}
\def\s{\shadow}
\usepackage{textcomp}
\def\BibTeX{{\rm B\kern-.05em{\sc i\kern-.025em b}\kern-.08em
    T\kern-.1667em\lower.7ex\hbox{E}\kern-.125emX}}
\begin{document}
\title{Multitask-Informed Prior for In-Context Learning on Tabular Data: Application to Steel Property Prediction}
\author{Dimitrios Sinodinos$^{1,2}$, Bahareh Nikpour$^{1,2}$, Jack Y. Wei$^{1,2}$, Sushant Sinha$^3$, Xiaoping Ma$^4$, \\Kashif Rehman$^4$, Stephen Yue$^3$, and Narges Armanfard$^{1,2}$
\thanks{$^1$: Dep. of Electrical and Computer Engineering, McGill University}
\thanks{$^2$: Mila – Québec AI Institute}
\thanks{$^3$: Dep. of Mining and Materials Engineering, McGill University}
\thanks{$^4$: Algoma Steel Inc.}}

\maketitle

\begin{abstract}
Accurate prediction of mechanical properties of steel during hot rolling processes, such as Thin Slab Direct Rolling (TSDR), remains challenging due to complex interactions among chemical compositions, processing parameters, and resultant microstructures. Traditional empirical and experimental methodologies, while effective, are often resource-intensive and lack adaptability to varied production conditions. Moreover, most existing approaches do not explicitly leverage the strong correlations among key mechanical properties, missing an opportunity to improve predictive accuracy through multitask learning. To address this, we present a multitask learning framework that injects multitask awareness into the prior of TabPFN—a transformer-based foundation model for in-context learning on tabular data—through novel fine-tuning strategies. Originally designed for single-target regression or classification, we augment TabPFN's prior with two complementary approaches: (i) target averaging, which provides a unified scalar signal compatible with TabPFN’s single-target architecture, and (ii) task-specific adapters, which introduce task-specific supervision during fine-tuning. These strategies jointly guide the model toward a multitask-informed prior that captures cross-property relationships among key mechanical metrics. Extensive experiments on an industrial TSDR dataset demonstrate that our multitask adaptations outperform classical machine learning methods and recent state-of-the-art tabular learning models across multiple evaluation metrics. Notably, our approach enhances both predictive accuracy and computational efficiency compared to task-specific fine-tuning, demonstrating that multitask-aware prior adaptation enables foundation models for tabular data to deliver scalable, rapid, and reliable deployment \s{in}for automated industrial \s{applications}quality control and process optimization in TSDR.
\end{abstract}

\begin{figure*}[htp!]  
    \centering
    \includegraphics[width=2\columnwidth]{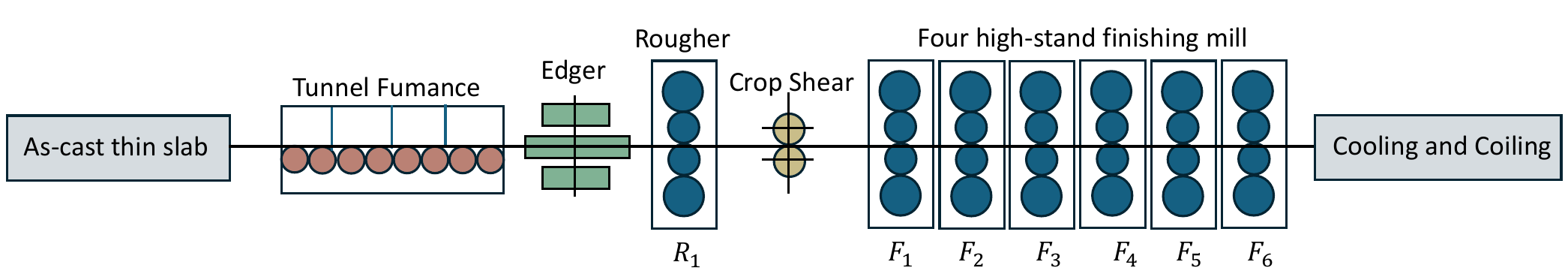} 
    \caption{Schematic of Direct Strip Processing Complex}
    \label{fig:overview}
\end{figure*}
\section{Introduction}
\label{sec:introduction}
\IEEEPARstart{S}{teel} is used widely across industries, including automotive \s{\cite{automotive}}, construction \s{\cite{construction}}, oil and gas \s{\cite{oil}}, and in various forms such as sheets, rods, and wires. The most critical step in forming steel into these shapes is the hot rolling (HR) process which is designed to both reduce the thickness of steel slabs and enhance their mechanical properties \cite{hotrolling}. Thin Slab Direct Rolling (TSDR) is a more advanced version of hot strip rolling that focuses on better productivity and energy efficiency \cite{tsdr}. The TSDR process starts by casting liquid steel into thin slabs about 70-80 mm thick. Right after casting, these slabs go into a tunnel furnace for a short heating of 15-20 minutes to ensure temperature homogeneity. Hence, the slab's width is adjusted with an edger, and then it is sent to the roughing stand. Here, the slab undergoes drastic reduction to form a transfer bar, about 35-40 mm thick. This bar moves on to the finishing stands where it's rolled in several subsequent stages to reach the desired thickness strip. Figure \ref{fig:overview} shows an overview of this process.

In the high-speed, high-temperature environment of a TSDR mill, process control is predominantly established before the rolling begins. To a process control engineer, the primary objective is to produce steel that not only meets specific thickness requirements but also achieves the desired mechanical properties, all while operating within the constraints of the rolling mill. The desired thickness of the HR steel is determined by the reduction settings at each rolling stand, which are configured before the start of the rolling process. The mechanical properties of the steel, on the other hand, depend on its chemical composition and the final microstructure, which is shaped by thermomechanical controlled processing (TMCP). TMCP involves careful adjustment of rolling speeds, reductions across different passes, and rolling temperatures, parameters that are crucial for tailoring the properties to meet specific demands. Given the complexity of the TSDR process, process control engineers must rely on predictive models that can estimate critical mechanical properties\s{, such as Lower Yield Strength (LYS), Ultimate Tensile Strength (UTS), and Elongation (ELO) measures,} in advance. These models enable fine-tuning of the rolling process and chemical composition, ensuring efficient, safe, and specification-compliant production while minimizing the need for costly post-production adjustments and testing. Accurate predictions require a robust understanding of thermomechanical processes and the intricate relationships between the material's chemical composition, microstructure, and processing parameters, which are essential for optimizing production and quality control. 

Early machine learning approaches to steel property prediction primarily relied on classical models such as support vector machines\s{\cite{wang2006application}} and gradient-boosted decision trees \cite{XGBoost}\s{\cite{takalo2022explainable}}. While effective in many scenarios, these methods struggle to capture complex feature dependencies and fail to generalize well in high-dimensional tabular datasets. More recently, deep learning models, such as feedforward neural networks (FNNs), recurrent architectures (RNNs, LSTMs), and convolutional neural networks (CNNs), have been explored for steel property prediction. Although these methods aim to extract meaningful representations from chemical and process data to enhance prediction accuracy\s{\cite{xie2021online}}, single-task deep learning models often struggle to capture the interdependencies between multiple mechanical properties, limiting their generalization in dynamic industrial settings. To address this limitation, multi-task learning (MTL) \cite{caruana1997multitask} has been increasingly adopted as a strategy to improve efficiency and generalization by jointly modeling multiple related tasks. MTL approaches range from hard parameter sharing, where a common feature representation is learned for all tasks, to soft parameter sharing, which allows task-specific representations while enabling knowledge transfer \cite{sinodinos2022attentive}. In addition to improving predictive performance, MTL offers notable computational benefits—particularly in scenarios involving limited data or costly training resources—by reducing the number of independent models that need to be trained and maintained \cite{sinodinos2025cross}. This shared architecture results in faster inference times and reduced memory usage, making MTL especially advantageous for real-time or large-scale industrial deployment.

Recently, Tabular Prior-data Fitted Network (TabPFN) \cite{hollmann2025accurate} has achieved state-of-the-art performance on small tabular datasets by using in-context learning (ICL), where a model adapts to new tasks directly from examples in its input context without requiring parameter updates. As a transformer-based Prior-Data Fitted Network (PFN), TabPFN is trained on synthetic datasets to approximate Bayesian posterior inference in a single forward pass, enabling rapid generalization across tasks. It represents a step toward foundation models for tabular data—models pre-trained to transfer broadly to downstream applications with minimal supervision. Recent work has shown that fine-tuning TabPFN’s prior on domain-specific datasets can further improve inference-time performance over a fully general prior~\cite{hollmann2025accurate}. While effective in single-task contexts, in MTL, this approach requires training a separate model for each target, which creates redundancy in compute and storage, and fails to exploit cross-task information that could improve generalization. Moreover, because TabPFN is pre-trained to predict a single target variable, performing multitask pre-training would require redesigning the output interface and generating new synthetic datasets that model joint target distributions—a major theoretical and computational undertaking. As a simpler and more practical alternative to predicting multiple continuous-valued mechanical properties, we propose injecting multitask awareness during fine-tuning by encoding the targets into a single signal, thereby preserving the off-the-shelf usability of pre-trained TabPFN models. We explore two strategies for implementing this idea: (1) fine-tuning on the average of all task targets, which provides a simple scalar proxy signal compatible with TabPFN’s single-target interface, and (2) incorporating a lightweight multitask adapter that introduces task-specific predictions during fine-tuning. It has also been shown with TabPFN \cite{hollmann2025accurate} that fine-tuning on related but not identical signals can outperform zero-shot inference, suggesting that proxy objectives can yield stronger priors. Building on this insight, we investigate whether similar benefits hold when fine-tuning on a joint multitask signal. Our results show not only improvements over zero-shot performance, but also, perhaps surprisingly, superior accuracy compared to fine-tuning on each exact task signal separately. By fine-tuning on multiple related tasks simultaneously, we not only consolidate all targets into a single model, but also leverage cross-task information to shape a more effective prior. This improves efficiency in terms of compute, time, and storage, while enhancing generalization. Our experiments on a dataset from a TSDR facility confirm these advantages, with multitask fine-tuning consistently surpassing its single-task counterpart. Together, these contributions demonstrate how foundation-model-style approaches for tabular data can be extended to complex multitask industrial applications that demand both high predictive performance and scalability.



\section{Related Works}
\subsection{Machine Learning for Steel Property Prediction}
The steel industry, a cornerstone of various sectors such as automotive, construction, and energy, has increasingly adopted machine learning (ML) techniques to address long-standing challenges associated with property prediction and process optimization \cite{shen2022multistep, sinha2024neural, sinha2025interpretable}. This adoption is enabled by the increasing availability of process data from modern rolling mills, which has encouraged the development of intelligent prediction and control systems that enhance operational decision-making. Prior studies have demonstrated the effectiveness of ML-based models in hot rolling for tasks such as predicting necking widths~\cite{9162543}, detecting surface defects~\cite{10413191}, forecasting temperatures~\cite{9944864}, and diagnosing faults in rolling mills~\cite{10644078}. In a similar direction, this work leverages chemical composition and process parameters to automatically predict key mechanical properties of steel, providing a data-driven tool to support quality control and process optimization.

Early applications of machine learning in the steel industry primarily utilized classical approaches like regression models, decision trees\s{\cite{takalo2022explainable}}, and support vector machines, which demonstrated success in tabular data analysis \cite{bessa2024least}. However, the advent of advanced machine learning methods, such as gradient-boosted decision trees (GBDTs) and neural networks, significantly enhanced predictive capabilities. GBDTs, in particular, gained prominence due to their ability to handle heterogeneous features and complex interactions while maintaining interpretability. These classical methods, although robust, are often limited in their capacity to fully exploit high-dimensional and non-linear relationships inherent in steel manufacturing processes.

Recent advances in deep learning, have further expanded the potential of machine learning in steel manufacturing. Xie et al. \cite{xie2021online} designed a deep neural network with tuned parameters to predict the mechanical properties of hot-rolled steel. Jiao et al. \cite{jiao2021remaining} enhanced the efficiency of hot continuous rolling by extracting both coarse-grained and fine-grained features from batch data and leveraging recurrent neural networks to predict the roll's remaining service life and health status. Rath et al. \cite{rath2020neural} integrated an adaptive algorithm with an artificial neural network to estimate the yield strength, ultimate tensile strength, and elongation of hot-rolled steel.\s{Dong et al.\cite{dong2020just} introduced a multi-block weighted semi-supervised soft sensor for predicting multiple mechanical properties of hot-rolled steel strips, utilizing the partial least squares method to construct low-correlation fusion sub-blocks for improved model performance.} Li et al. \cite{li2023deep} developed a multi-grain and multi-level forest framework that incorporates conventional process parameters with multi-grain scanning information to enhance prediction accuracy in strip processing.\s{ Xu et al. \cite{xu2019mechanical} employed convolutional neural networks, leveraging local connectivity and weight sharing to reduce model complexity, while using convolutional operations to extract key feature information for improved mechanical property predictions.}\s{In the context of mutli-task learning, Yang et al. [17] employed a dynamic task balancing strategy to improve multi-task learning performance, while Zeng et al. [18] utilized MTL to address data imbalance and enhance the predictive accuracy of multiple material properties. Yan et al. [19] demonstrated that MTL can be used to jointly predict mechanical responses, crack morphology, and stress-strain relationships in composite materials, highlighting its ability to capture cross-task dependencies. Despite these advancements, existing multi-task learning approaches often rely on either rigid parameter sharing or naive task interactions, which can lead to suboptimal generalization when modeling complex physical processes.}
\subsection{Deep Learning with Tabular Data}
Tabular data refers to data organized as rows representing samples, and columns representing features. In steel property prediction, tabular datasets are structured such that each row corresponds to a steel batch and columns represent chemical compositions, processing parameters, and mechanical properties. Unlike unstructured data such as images, text, or audio, which typically consist of homogeneous feature types, tabular or structured data are defined by their heterogeneous feature types. These may include numerical, categorical, and textual attributes with complex interdependencies. The history of machine learning for tabular data dates back several decades, where classical methods like linear and logistic regression\s{\cite{lin_reg}}, decision trees\s{\cite{dts}}, and support vector machines (SVMs)\s{\cite{svm}} have been the standard for tabular classification and regression tasks. Among classical methods, Gradient Boosted Decision Trees (GBDT) \cite{XGBoost}, based on ensembles of decision trees, demonstrated great success because of their interpretability and proficiency in handling heterogeneous features and intricate feature interactions.

In recent years, deep learning models have gained popularity by offering the potential for improved performance, especially for complex, high-dimensional, and large-sized dataset. Despite the success of deep learning on images, natural languages and speech data, standard neural network architectures such as multi-layer perceptrons (MLPs)\s{\cite{mlp}}, convolutional neural networks (CNNs), and recurrent neural networks (RNNs) \s{\cite{lecun2015deep}} often underperform in the tabular domain when compared to GBDTs, while also facing drawbacks in efficiency and explainability. Current research in deep learning for tabular data seeks to address the shortcomings of deep neural networks\s{ by exploring novel approaches}\cite{Net_DNF}. \s{One trend involves replicating the success of ensembles of decision tree with deep learning. Differentiable decision trees such as NODE \cite{NODE} and Net-DNF \cite{Net_DNF} aim to harness the flexibility of neural networks while maintaining the performance and interpretability of decision trees. Another line of work focuses on deep regularization techniques, Shavitt \& Segal \cite{shavitt2018regularization} and Kadra et al. \cite{regularization_mlp} demonstrated that simple neural networks on tabular data can be competitive by adding special constraints to the learned weights. Finally, }Inspired by the success of transformers in natural language processing, an array of recent models leverage attention mechanisms to capture intricate relationships among features or samples and introduce explainability to black-box neural networks through attention maps. Huang et al. \cite{tab_transformer} proposed TabTransformer which first transforms categorical features into contextual embeddings which are subsequently passed into fully connected layers with numerical features. SAINT \cite{saint} improves this approach by embedding both categorical and numerical features as inputs to a transformer and leverages inter-sample attention to improve performance and robustness to noisy data. \s{Finally, as a forerunner to deep learning models, TabNet \cite{tabnet} \blue{DS: Should we include this if we don't test with it?} employs a sequential multi-step architecture analogous to sequential decision steps in tree-based methods. At each decision step, the model performs instance-wise feature selection and passes selected features into a self-attention transformer. The model then makes inference based on outputs of each decision step and provides hierarchical explainability based on selected features. }

Machine learning on tabular data finds applications in various industries. The promising applications domains include, but are not limited to, clinical diagnostics\s{\cite{ML_clinical}}, insurance\s{\cite{ML_insurance}}, genomics\s{\cite{ML_genomics}} and manufacturing \cite{ML_industrial}.\s{While current applications of ML on tabular data primarily employ traditional non-deep learning models. Numerous recent studies have investigated the use of TabNet for insurance pricing \cite{tab_net_insurance}, vehicle-pedestrian crash prediction \cite{tabnet_pedestrian} and classification of irregularities on roads \cite{tabnet_road}.} Despite recent advances in machine learning for tabular data, several challenges remain. First, missing features are common in real-world tabular datasets, necessitating models that can handle incomplete inputs without compromising performance and remain robust to noisy data. Second, many practical applications involve limited data availability, highlighting the need for models with high sample efficiency that can learn effectively from small datasets. Lastly, while some deep learning models offer local interpretability through attention maps or feature selection masks, it is equally important to capture and communicate global reasoning processes to improve overall model explainability. Addressing these challenges is essential for broadening the practical adoption of machine learning in tabular domains. \s{we can merge TabPFN with this section later}


\subsection{MultiTask Learning}
In recent years, multitask learning (MTL) \cite{caruana1997multitask} has emerged as a parameter-efficient learning paradigm that can also outperform traditional single-task learning (STL). Generally, MTL involves a single network that can learn multiple tasks by jointly optimizing multiple loss functions. Consequently, employing a single network implies the sharing of several layers or features between tasks. In numerous dense prediction cases, sharing features across tasks has been demonstrated to enhance per-task performance while utilizing fewer per-task model parameters. This improvement is attributed to enhanced generalization by leveraging domain-specific knowledge between related tasks \cite{sinodinos2022attentive}. The prominent research directions for modern MTL methods explore either the optimization strategy\s{\cite{xin2022current}} or the design of the deep multitask architecture \cite{sinodinos2022attentive, sinodinos2025cross, sinodinos2026multitab}.\s{Historically, deep architecture design can be further categorized based on the parameter-sharing scheme. Specifically, multitask models can employ hard-parameter sharing or soft-parameter sharing \cite{9336293}. In hard-parameter sharing, there is a portion of the network (typically the encoder) that is shared among all tasks, and there are task-specific layers (typically decoders) unique to each task. In soft-parameter sharing, each task has their own set of parameters, while another set of parameters is dedicated to sharing features across tasks (i.e., cross-talk).} 

In the application of multitask prediction for material properties, several approaches have been proposed to address domain-specific challenges. For instance, Yang et al. \cite{yang2022multitask} introduced a dynamic task-balancing strategy that accounts for task differences and complexities, reducing overfitting across multiple quantum mechanical properties.\s{ Zeng et al. \cite{zeng2021multi} applied MTL to address unbalanced data and multi-objective prediction in negative thermal expansion materials, predicting initial temperature, end temperature, and negative expansion coefficients.}\s{Similarly, Yan et al. [19] modeled the transverse mechanical response, crack morphology, and stress-strain behavior in carbon fiber-reinforced polymers, demonstrating strong generalization capabilities.} Ma et al. \cite{ma2018modeling} proposed a Multi-gate Mixture of Experts (MMoE) model composed of multiple MLPs to balance shared and task-specific representations. More recently, a Multi-gate Mixture of Convolutional Experts (MMCE) \cite{zhang2024multi} framework has been introduced for hot-rolled steel property prediction, incorporating convolutional attention to extract localized process features and improve interpretability.

Despite these advancements, several challenges remain. Many existing MTL architectures rely on rigid parameter sharing, which limits their adaptability to nuanced relationships between tasks. In tabular domains like steel manufacturing, where features are heterogeneous and lack spatial structure, convolutional modules may not fully capture the underlying global dependencies. Moreover, many of these models involve complex architectures that require significant training time and hyperparameter tuning, making them less suitable for real-time or rapidly evolving production settings. These limitations motivate the need for lightweight, flexible, and efficient multitask models that generalize well under limited data and operational constraints.

\begin{figure*}[htp!]  
    \centering
    \includegraphics[width=2\columnwidth]{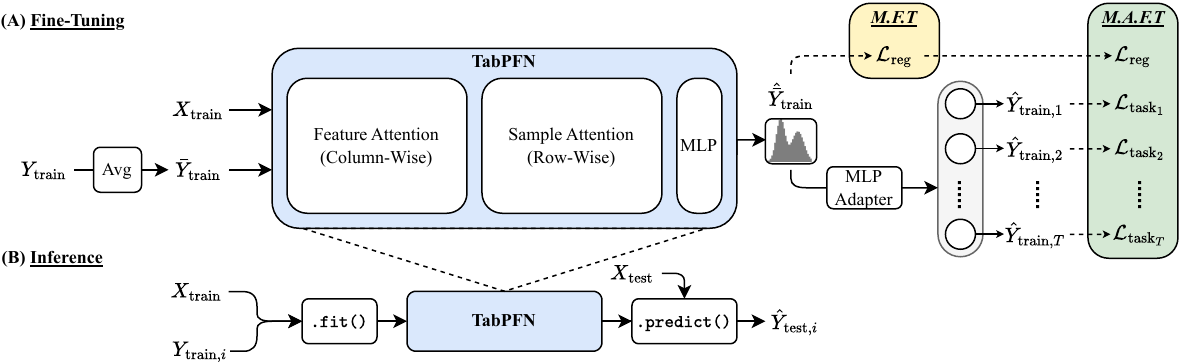} 
    \caption{An overview of our proposed multitask fine-tuning techniques and the standard TabPFN inference pipeline. \textbf{(A) Fine-tuning}: $X_{\text{train}} \in \mathbb{R}^{N_\text{train},D}$ and $Y_{\text{train}} \in \mathbb{R}^{N_{\text{train}},T}$ represent the train data and train labels respectively, where $N_{\text{train}}$ is the number of training samples, $D$ is the number of features per sample, and $T$ is the number of tasks. To obtain $\bar{Y}_{\text{train}}\in \mathbb{R}^{N_{\text{train}},1}$, we average (Avg) $Y_{\text{train}}$ along the task dimension. Using $X_{\text{train}}$ and $\bar{Y}_{\text{train}}$ as inputs to the underlying transformer of TabPFN, we obtain the predicted $\hat{\bar{Y}}_{\text{train}}$. In the standard multitask fine-tuning (M.F.T), we use only the regression loss $\mathcal{L}_{\text{reg}}$ as the training signal. In multitask adapter fine-tuning (M.A.F.T), we also use the task-specific losses after obtaining the task-specific predictions $\hat{Y}_{\text{train},i}\in \mathbb{R}^{N_{\text{train}},1}$ for $i\in[1,T]$ by passing $\hat{\bar{Y}}_{\text{train}}$ through an MLP adapter. \textbf{(B) Inference}: We sequentially predict the $i$th task target by following the standard TabPFN inference pipeline. This involves first updating the prior statistics of our pre-trained TabPFN using the $\texttt{.fit()}$ function with $X_{\text{train}}$ and $Y_{\text{train},i}$ as inputs, followed by a forward pass using $\texttt{.predict()}$ with $X_{\text{test}}\in \mathbb{R}^{N_{\text{test}},D}$ as input to obtain $\hat{Y}_{\text{test},i}\in \mathbb{R}^{N_{\text{test}},1}$, which is the target for the $i$th task and $N_{\text{test}}$ is the number of test samples.}
    \label{fig:bd}
\end{figure*}

\section{Methodology}
We leverage TabPFN, a transformer-based foundation model designed for small-sized tabular datasets, as a base learner for predicting multiple correlated mechanical properties of steel. While TabPFN was originally built for single-target prediction, we extend its capabilities through two multitask fine-tuning strategies, while preserving its in-context learning interface at inference. A central challenge is that TabPFN is pre-trained with a single target column; extending it directly to multitask learning would require architectural redesign and regeneration of synthetic pretraining datasets that capture joint target distributions, which is a major theoretical and practical challenge. Instead, we adopt lightweight fine-tuning methods that encode multitask information into forms that remain compatible with TabPFN’s single-target design. The first approach fine-tunes the model using the average of all task targets as a single, unified training signal. The second augments the model with a lightweight multitask adapter module that enables task-specific supervision during fine-tuning. Both strategies improve computational efficiency by allowing one model to be fine-tuned and stored across all tasks, whereas a single-task approach would require training and maintaining a separate model for each property. In the following sections, we first provide an overview of TabPFN’s architecture and then describe our multitask extensions in detail.

\subsection{TabPFN Overview}

TabPFN is a prior-fitted network designed for tabular data that approximates Bayesian inference via in-context learning \cite{hollmann2025accurate},  tailored for tabular classification and regression tasks. \s{Unlike traditional machine learning pipelines, which typically require dataset-specific training and hyperparameter tuning to achieve strong performance, prior fitted networks (PFNs) adopt a more generalizable paradigm. In PFNs, the network is first trained on a large set of datasets to learn a prior over the target tasks. Then, at inference time, a PFN can infer a posterior predictive distribution on arbitrary datasets in a single forward propagation.}Unlike traditional deep learning models, TabPFN does not require any additional training when presented with a new tabular dataset. Instead, it directly takes a set of labeled training samples along with test samples and produces class probability predictions in a single forward pass, without performing any gradient-based updates. This behavior mirrors that of Gaussian Processes and reflects the ICL capabilities often seen in large language models.

\s{TabPFN trains its prior on synthetic datasets with varying causal structures and dataset properties aligned with real-world data. After conditioning the model on a training set, TabPFN can then perform classification and regression tasks on one target of the test set, without additional gradient updates. This enables fast adaptation on moderate-sized tabular datasets (i.e., fewer than 10,000 samples) by bypassing the need for per-task training and hyperparameter tuning. Moreover, compared to traditional ML models, TabPFN is inherently more robust to overfitting because of its approximation of Bayesian inference. Instead of selecting a single model that best fits the training data, TabPFN approximates the posterior predictive distribution, and effectively integrates over many plausible models conditioned on the training set. This Bayesian marginalization discourages overconfident predictions and reduces the risk of fitting noise or spurious patterns in small datasets. \s{Empirical results show that TabPFN outperforms strong tabular baselines trained on the target dataset with significant speedups. Our experiments further corroborate these findings.}}

To better leverage the tabular structure, TabPFN encodes both features and labels as tokens and applies a two-way attention mechanism: first, across features within each sample (column-wise), and then across all samples for each feature (row-wise). This enables the model to be invariant to both feature and sample order, improving generalization and scalability. The model is trained offline on a large distribution of synthetic classification tasks sampled from a probabilistic prior, allowing it to internalize robust inductive biases that generalize well to real-world datasets. Its flexible token-based architecture allows it to attend jointly to the training and test data, making it particularly suitable for small-sized tabular problems that demand fast, accurate, and adaptive predictions.

In addition to classification, TabPFN supports regression through a discretized output representation known as the support bar distribution. For a single sample $(x, y)$ with continuous target $y \in \mathbb{R}$, the target is mapped to a probability distribution $\mathbf{p}(y) \in \Delta^{K-1}$ over a fixed support of $K$ bins with centers $\{c_k\}_{k=1}^{K}$. Given model logits $\mathbf{z}(x)$, the predicted probabilities are
\begin{equation}
    q_k(x) = \mathrm{softmax}(\mathbf{z}(x))_k, \quad k=1,\dots,K .
\end{equation}

The regression training loss is defined as the cross-entropy between the soft label distribution $\mathbf{p}(y)$ and the model prediction $\mathbf{q}(x)$:
\begin{equation}
    \mathcal{L}_{\text{reg}}(x,y) = -\sum_{k=1}^{K} p_k(y) \,\log q_k(x) .
\end{equation}

At inference time, the scalar prediction for a sample is given by the expectation over the support,
\begin{equation}
    \hat{y}(x) = \sum_{k=1}^{K} q_k(x)\,c_k .
\end{equation}
For a training set $(X_{\text{train}}, Y_{\text{train}})$ used in fine-tuning, we denote by $\hat{Y}_{\text{train}}$ the collection of predicted support bar outputs $\{\hat{y}(x)\}_{x \in X_{\text{train}}}$.
This dataset-level notation naturally extends to the averaged targets $\bar{Y}_{\text{train}}$ and their predictions $\hat{\bar{Y}}_{\text{train}}$, and to the task-specific predictions $\hat{Y}_{\text{train},i}$ introduced in the subsequent sections.

\subsection{Multitask-Informed Prior Adaptation}
While TabPFN demonstrates strong zero-shot capabilities, domain-specific fine-tuning is known to further enhance task-specific performance. However, in a multitask setting, single-task fine-tuning is inefficient and fails to capture cross-task knowledge. Furthermore, performing multitask fine-tuning is challenged by TabPFN's architectural restriction to a single predictive output, which renders it structurally incapable of modeling the full joint posterior predictive distribution $p(y_1, \dots, y_T \mid x)$ without significant architectural changes and costly retraining. To overcome this limitation, we propose a unified fine-tuning framework that injects multitask knowledge via an optimal ``proxy'' signal. Building on findings that fine-tuning on proxy targets can outperform zero-shot performance~\cite{hollmann2025accurate}, we posit that the most effective multitask proxy is the mathematical summary of the joint distribution. Thus, we formulate the adaptation process as a constrained multitask optimization problem, which involves identifying the Maximum Likelihood Estimator (MLE) for the shared latent structure underlying the tasks.

\subsubsection{Joint Signal Injection via Target Averaging}

To derive this optimal proxy, we adopt the Bayesian multitask formulation of Kendall et al. \cite{kendall2018multi}, which states that task correlations are mediated by a shared latent sufficient statistic $z$, subject to task-specific aleatoric uncertainty. Unlike \cite{kendall2018multi}, where task-specific noise parameters ($\sigma_i$) are learned simultaneously with the model, we require a fixed aggregation target for the foundation model's prior \textit{before} task-specific parameters are learned. Allowing the prior to learn variable task weights dynamically creates a trivial shortcut where the optimization collapses the shared target to the single task with the lowest noise. To prevent this, we apply the \textbf{principle of indifference}, imposing a \textbf{Uniform Homoscedasticity} assumption across all tasks ($\sigma_i = \sigma$). 

We model the residual task-specific variation as independent Gaussian noise with a shared variance $\sigma^2$:
\begin{equation}
    y_i = z + \epsilon_i \quad \text{where} \quad \epsilon_i \sim \mathcal{N}(0, \sigma^2)
\end{equation}
This implies the conditional likelihood for each task is given by the Gaussian probability density function:
\begin{equation}
    p(y_i \mid z) = \frac{1}{\sqrt{2\pi\sigma^2}} \exp\left( -\frac{(y_i - z)^2}{2\sigma^2} \right)
\end{equation}
The ideal multitask objective is to find the estimator $z^*$ that maximizes the joint likelihood of observing all tasks simultaneously:
\begin{equation}
    z^* = \mathop{\mathrm{argmax}}_{z} p(y_1, \dots, y_T \mid z)
\end{equation}
Due to the conditional independence (implied by $z$ being the sufficient statistic), the joint likelihood factorizes into the product of individual task likelihoods. Maximizing this joint likelihood is equivalent to minimizing the Negative Log-Likelihood (NLL):
\begin{align}
    \mathcal{L}_{NLL}(z) &= -\log \left( \prod_{i=1}^{T} p(y_i \mid z) \right)
\end{align}
Substituting the Gaussian density function defined above, the log-likelihood expansion becomes:
\begin{align}
    \mathcal{L}_{NLL}(z) &= - \sum_{i=1}^{T} \left( \log \frac{1}{\sqrt{2\pi\sigma^2}} - \frac{(y_i - z)^2}{2\sigma^2} \right)
\end{align}
Since $\sigma$ is constant across tasks (per our uniform assumption), minimizing $\mathcal{L}_{NLL}$ is proportional to minimizing the unweighted sum of squared errors $J(z)$:
\begin{equation}
    \mathop{\mathrm{argmin}}_{z} \mathcal{L}_{NLL}(z) \equiv \mathop{\mathrm{argmin}}_{z} \sum_{i=1}^{T} (y_i - z)^2
\end{equation}
To find the optimal latent estimator $z^*$, we set the first derivative of $J(z)$ to zero:
\begin{equation}
    \frac{\partial J}{\partial z} = -2 \sum_{i=1}^{T} (y_i - z) = 0 \implies z^* = \frac{1}{T} \sum_{i=1}^{T} y_i = \bar{y}
\end{equation}
To confirm this critical point is a global minimum, we verify the second derivative $\frac{\partial^2 J}{\partial z^2} = 2T > 0$, confirming convexity.

We can now conclude that the arithmetic mean $\bar{y}$ is the unique MLE for the sufficient statistic $z$ under the uniform homoscedastic assumption. Therefore, by fine-tuning TabPFN to predict $\bar{y}$ (using its native cross-entropy loss \cite{hollmann2025accurate}), we are training the foundation model to learn the optimal shared representation that best explains the joint task distribution under the single-head constraint. At inference time, the model uses this learned ``Centroid Prior'' as a robust starting point. The ICL interface then uses the provided support set $(X_{support}, Y_{task_i})$ to shift the probability mass from the shared centroid $z^*$ to the specific task posterior required.

\subsubsection{Task-Specific Disentanglement via Adapter Supervision}

While the Centroid Prior optimally captures the shared signal $z^*$ (the mean), real-world scenarios typically match the full \textit{Task-Dependent Homoscedastic} setting described by \cite{kendall2018multi}, where tasks exhibit varying noise scales ($\sigma_i \neq \sigma_j$) or negative correlations (e.g., strength vs. ductility).

To address this, the second step augments the framework with a lightweight Multilayer Perceptron (MLP) adapter during fine-tuning. This allows the system to recover the flexibility to model task-specific heteroscedasticity and non-linear projections which were simplified during the prior adaptation phase. For each task $i$, the adapter produces a scalar prediction $\hat{Y}_{\text{train},i}$ defined as a transformation of the averaged prediction $\hat{\bar{Y}}_{\text{train}}$ through the MLP adapter:
\begin{equation}
    \hat{Y}_{\text{train},i} = f_{\text{adapter},i}\big(\hat{\bar{Y}}_{\text{train}}\big),
\end{equation}
where $f_{\text{adapter},i}$ is the $i$-th output head of the adapter. The per-task loss for task $i$ is then given by the mean squared error between the adapter prediction and the true label,
\begin{equation}
    \mathcal{L}_{\text{task}_i} = \frac{1}{N}\sum_{n=1}^{N}\big(\hat{Y}_{\text{train},i}^{(n)} - Y_{\text{train},i}^{(n)}\big)^2 ,
\end{equation}
where $N$ is the number of training samples. The overall training objective combines the primary TabPFN regression loss $\mathcal{L}_{\text{reg}}$ with the auxiliary task-specific losses:
\begin{equation}
    \mathcal{L}_{Total} = \mathcal{L}_{reg} + \lambda \frac{1}{T} \sum_{i=1}^{T} \mathcal{L}_{task_i}
\end{equation}
Here, $\mathcal{L}_{reg}$ anchors the foundation model to the shared MLE centroid, while the adapter heads, optimized via $\mathcal{L}_{task_i}$, provide the necessary degrees of freedom to model task-specific heteroscedasticity. Crucially, the adapter is removed at inference, preserving the model's efficient single-pass interface while retaining the disentangled representation in the updated prior.

\section{Experimental Setup}

\begin{table*}[hbt!]
\centering
\caption{Summary of Features and Targets}
\label{tab:features_targets_summary}
\begin{tabularx}{\textwidth}{|l|X|}
\hline
\rowcolor[gray]{0.9} \textbf{Feature Category} & \textbf{Features} \\
\hline
\textbf{Chemical Composition} & Al, B, C, Ca, Cr, Cu, Mn, Mo, Nb, Ni, P, S, Si, Sn, Ti, V, N \\
\hline
\textbf{Caster Section} & Avg. Casting Speed, Superheat \\
\hline
\textbf{Slab Geometry} & Slab Width, Slab Gauge, Width (R1 Exit) \\
\hline
\textbf{Strip Geometry} & Final Width \\
\hline
\textbf{Thickness Measurements} & Exit Thickness (R1, F1, F2, F3, F4, F5, F6) \\
\hline
\textbf{Furnace Parameters} & Slab Residence Time \\
\hline
\textbf{Rolling Speed} & Roll Speed (R1, F1, F2, F3, F4, F5, F6) \\
\hline
\textbf{Roll Forces} & Roll Force (R1, F1, F2, F3, F4, F5, F6) \\
\hline
\textbf{Temperature Measurements} & R1 Entry Temperature, FM Entry Temperature, Finishing Temperature, Coiling Temperature \\
\hline
\hline 
\rowcolor[gray]{0.9} \textbf{Target Category} & \textbf{Targets} \\
\hline
\textbf{Mechanical Properties} & Lower Yield Strength (MPa), 0.2\% Offset Yield Strength (MPa), 0.5\% Extension Yield Strength (MPa), Ultimate Tensile Strength (MPa), 2" Elongation \\
\hline
\end{tabularx}
\end{table*}

\subsection{Dataset}

The dataset used in this study was obtained from the Thin Slab Direct Rolling (TSDR) facility at a steel production company.\s{Algoma Steel Inc., located in Ontario, Canada.} After cleaning the raw data, we retained a total of 6,415 instances of V-based micro-alloyed steel. To facilitate model development and validation, the dataset was split into a 70:30 ratio, with 70\% allocated for training and 30\% reserved for testing.

The final properties of hot-rolled (HR) steel are influenced by both its chemical composition and its processing history. In this dataset, we included a total of 49 features, which encompass 17 chemical elements and 32 processing parameters. Lack of longer reheating times in the TSDR process makes the upstream casting process influential, hence, average casting speed and superheat were included. Additionally, features related to the geometry of the slab and final strip were also incorporated. From the homogenization furnace, we included the discharge temperature and the residence time of the slabs. In terms of physical metallurgy, TMCP in the rolling mill is characterized by three main parameters: strain, strain rate, and temperature at each stage. The thickness and roll speed at each stand provide information on the strain and strain rate during rolling. Although temperature is a critical factor, data was only available for four temperatures: the entry temperature at the roughing and finishing stands, the finishing temperature, and the coiling temperature. Given the importance of temperature, and the fact that it was not available at every stand, we also included roll force data at each stand. While roll force might not initially seem like a direct causal factor, it provides valuable information into the material's flow behavior at each stage, which is inherently linked to temperature. The complete list of features and targets is summarized in Table \ref{tab:features_targets_summary}.

\subsection{Tasks Overview}\label{Tasks Overview}
Given the wide range of potential customer applications, it is impractical to test the HR steel’s performance for every possible use case. Therefore, a standardized testing protocol is established to evaluate the material's properties, which can then be used by end users to select the appropriate steel for their needs. A commonly employed test is the uniaxial tensile test, where a sample of the HR strip is prepared in a standard rectangular geometry (dogbone shape). During the test, the material is subjected to a uniaxial tensile load applied at its ends, and its behavior is observed as the load increases. The result is a stress-strain curve, from which a number of critical points are extracted. Commonly extracted metrics include Lower Yield Strength (LYS), 0.2\% Offset Yield Strength (OYS), 0.5\% Extension Yield Strength (EYS), Ultimate Tensile Strength (UTS), and two-inch Elongation (ELO). These are the key property metrics that we have aimed to predict in this work. Each of these properties provides crucial information about the material’s performance and suitability for specific applications. LYS indicates the stress at which the material begins to deform plastically. This is essential for applications requiring materials that can withstand significant stress without permanent deformation. The OYS and EYS offer insights into the material's behavior under small, specified strains, often useful when the material does not have a pronounced yield point. UTS represents the maximum stress that the material can withstand before failure, making it a critical factor in ensuring the material can handle extreme conditions without fracture. Finally, ELO measures the ductility of the material, indicating how much it can deform before fracture, which is key for applications requiring toughness. Often, engineers assess these properties in combination to determine the material's applicability.\s{For example, to qualify as V-based HSLA ASTM A572 Grade 50, a common structural material, the steel must have a minimum YS of 50 Ksi, tensile strength of 65 Ksi, and 2" elongation of 21\%.}

\begin{figure}[htb!]
    \centering
    \includegraphics[width=\linewidth]{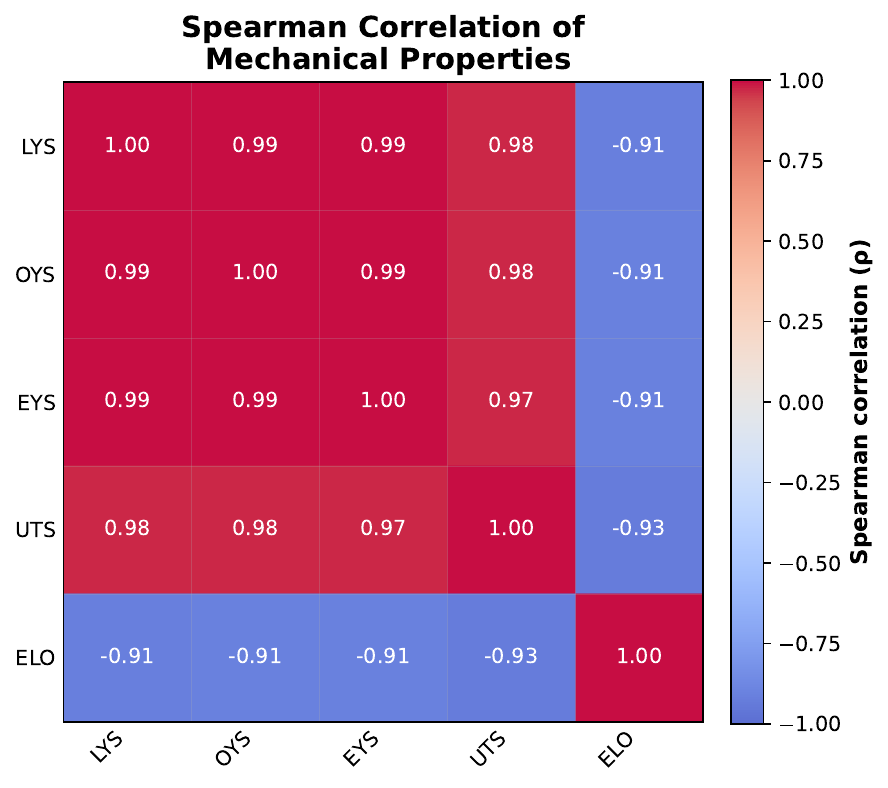}
    \caption{Spearman correlation matrix of the five mechanical property targets (LYS, OYS, EYS, UTS, ELO). Values represent pairwise rank correlations, with positive correlations shown in red and negative correlations in blue.}
    \label{fig:spearman}
\end{figure}

To better understand the relationships among these mechanical properties, we computed a Spearman rank-correlation matrix and visualized it as a heatmap in Figure~\ref{fig:spearman}. Spearman correlation quantifies how strongly pairs of variables vary together in a monotonic fashion, independent of exact scale or linearity. The resulting matrix reveals that LYS, OYS, EYS, and UTS are all highly and positively correlated ($\rho \approx 0.97$--$1.00$), reflecting their shared physical basis in the onset and progression of plastic deformation. In contrast, ELO shows a strong negative correlation ($\rho \approx -0.9$) with all four strength measures, consistent with the well-known strength--ductility trade-off in steel where increasing strength typically reduces ductility. These pronounced correlations suggest that cross-task information, particularly the commonalities among the strength metrics and their systematic opposition to elongation, can be leveraged in a multitask learning framework to improve predictive performance beyond what is possible when tasks are modeled in isolation.

\subsection{Performance Metrics}
To assess model performance across different mechanical properties, we employ the following four metrics:

\textbf{Mean Absolute Error (MAE\%)} quantifies the average magnitude of prediction errors relative to the true values, expressed as a percentage:

\begin{equation}
    \text{MAE\%} = \frac{1}{n} \sum_{i=1}^{n} \left|\frac{y_i - \hat{y}_i}{y_i}\right| \times 100
\label{eq:mae}
\end{equation}
where \( y_i \) is the true value, \( \hat{y}_i \) is the predicted value, and \( n \) is the number of samples. Lower MAE\% indicates better predictive accuracy.

\textbf{Predictive Accuracy Metric (PAM)} is used in this study as a key quantification metric for evaluating prediction performance. It is defined as the percentage of predictions that fall within a specified relative error threshold \(\epsilon\). PAM is computed as:

\begin{equation}
    \text{PAM} = \frac{NCP}{NP} \times 100
\label{eq:pam}
\end{equation}
where \(NP\) is the total number of predictions, and \(NCP\) is the number of correct predictions. A prediction is considered correct if the relative error with respect to the predicted value satisfies:

\begin{equation}
    \text{NCP} = \left| \frac{y_j - \hat{y}_j}{\hat{y}_j} \right| \leq \epsilon
\label{eq:ncp}
\end{equation}
where \( y_j \) is the true (measured) value and \( \hat{y}_j \) is the predicted value. In this work, we report PAM at \(\epsilon = 5\%\) and \(\epsilon = 2.5\%\), referred to as PAM-5 and PAM-2.5 respectively. Any prediction error beyond these thresholds is considered incorrect and reduces the overall PAM score.

\textbf{Explained Variance (EV)} measures the proportion of variance in the target values that is captured by the model predictions. It is defined as:

\begin{equation}
    \text{EV} = 1 - \frac{\text{Var}(y - \hat{y})}{\text{Var}(y)}
\label{eq:ev}
\end{equation}
where \( \text{Var}() \) denotes variance, and higher EV values suggest that the model better explains the variability in the data.

\textbf{MTL Gain}, as proposed by \cite{maninis2019attentive}, provides a comprehensive assessment of the performance improvement achieved by a multitask learning (MTL) method \(m\) compared to a single task learning (STL) baseline \(b\) across multiple tasks \(i \in T\). This assessment is encapsulated by Equation \ref{deltam}.

\begin{equation}
    \Delta_m = \frac{1}{T}\sum_i^T (-1)^{l_i} \left( \frac{M_{m,i} - M_{b,i}}{M_{b,i}} \right)
\label{deltam}
\end{equation}
In this equation, \(M_{m,i}\) and \(M_{b,i}\) represent the performance metric (MAE in our case) achieved by method \(m\) and baseline \(b\) respectively for task \(i\). The value \(l_i\) is set to 1 if a lower metric value is desirable for task \(i\), and 0 otherwise. The resulting \(\Delta_m\) is interpreted as a percentage improvement over the baseline \(b\) across all tasks. By employing MTL Gain, we can effectively gauge the holistic enhancement in predictive performance offered by the MTL approach compared to traditional STL methods.

\s{\textbf{Mean Absolute Error (MAE)} percentage is a commonly used metric to evaluate the accuracy of regression models. It measures the average absolute difference between the predicted and actual values as a percentage of the actual value, as seen in Eq. \ref{eq:mae}.

\begin{equation}
    \text{MAE\%} = \frac{1}{n} \sum_{i=1}^{n} \left|\frac{y_i - \hat{y}_i}{y_i}\right| \times 100
\label{eq:mae}
\end{equation}

where \( y_i \) is the actual value, \( \hat{y}_i \) is the predicted value, and \( n \) is the number of samples. Using MAE\%, we will also report the Percentile Absolute Error (PAM) at 5\% and 2.5\% thresholds as PAM-5 and PAM-2.5 respectively.  

\textbf{Variance Accounted For (VAF)} measures the proportion of variance in the target variable that is explained by the regression model. It is calculated by comparing the variance of the predicted values to the variance of the actual values. VAF provides insight into how well the model captures the variability in the data.

\begin{equation}
    \text{VAF} = 1 - \frac{\text{Var}(y - \hat{y})}{\text{Var}(y)}
\label{eq:vaf}
\end{equation}

where \( y \) is the vector of actual values, \( \hat{y} \) is the vector of predicted values, and \( \text{Var}() \) represents the variance operator, where the formula for the variance of a variable \( X \) is calculated as follows:

\begin{equation}
   \text{Var}(X) = \frac{1}{n} \sum_{i=1}^{n} (x_i - \mu)^2 
\end{equation}}

\subsection{Baselines} \label{baselines}
In our experiments, we evaluated our method in comparison to several well-established baseline approaches, each of which is described in detail below:
\begin{itemize}
  \item \textbf{STL (Single Task Learning)\s{\cite{goodfellow2016deep}}}: 
Single Task Learning trains a separate MLP model for each task independently. This approach does not share parameters or representations across tasks, which can limit its ability to leverage commonalities between tasks. STL often serves as a straightforward baseline for evaluating the effectiveness of multi-task models.
  \item \textbf{MTL (Multi-Task Learning) \cite{caruana1997multitask}}:
Multi-Task Learning jointly trains multiple tasks using shared representations, typically improving performance by leveraging commonalities across tasks. The implementation involves sharing a portion of the first layers in a MLP across all tasks, while the final output heads remain task-specific. By learning tasks in parallel, MTL can lead to better generalization than training each task in isolation, as tasks can reinforce shared underlying patterns in the data.

  \item \textbf{MMoE (Multi-gate Mixture of Experts) \cite{ma2018modeling}}:
The MMoE model is a specialized multi-task architecture that uses multiple “expert” sub-networks and separate gating networks for each task. The gating networks learn how to combine the outputs of these experts for each task, enabling both task-specific customization and efficient parameter sharing.
  \item \textbf{PLE (Progressive Layered Extraction) \cite{tang2020progressive}}: 
PLE extends the idea of Mixture-of-Experts by using a progressive extraction process to separate task-shared and task-specific features at different layers. By progressively distilling shared knowledge and allocating specialized layers to each task, PLE helps alleviate task interference and improves the overall multi-task performance.
  \item \textbf{STEM \cite{su2024stem}}: STEM is a multi-task recommendation framework designed to harness the power of embeddings for multiple objectives. It leverages shared embedding layers for feature extraction while incorporating additional task-specific embeddings to handle the nuances of different tasks.
  \item \textbf{FT-Transformer\cite{gorishniy2021revisiting}}: This method applies the Transformer architecture, originally developed for sequence data, to tabular data. By using attention-based mechanisms, it captures complex interactions among features. Its flexibility in handling categorical and numerical features makes it a strong baseline for many tabular tasks.
  \item \textbf{SAINT \cite{saint}}: SAINT is a specialized transformer model for tabular data that integrates row-wise attention and contrastive pre-training. This design helps the network learn inter-row relationships (similar to sequences) while also benefiting from a pre-training phase, often leading to improved generalization and data efficiency. 
   \s{ \item \textbf{MLP \cite{goodfellow2016deep}:} A Multilayer Perceptron (MLP) with one hidden layer comprises an input layer, a hidden layer, and an output layer. The hidden layer applies weighted sums of inputs followed by an activation function, enabling the model to learn complex relationships between inputs and outputs. This architecture, though simpler than deeper networks, can still effectively approximate various functions and solve a wide range of problems; making it a suitable single-task learning baseline model for our experiments.
    \item \textbf{AttResNet \cite{choi2020attentional}:} The Attentional Residual Network (AttResNet) consists of a residual network augmented with an attention mechanism. This architecture integrates residual connections for facilitating gradient flow and an attention mechanism for focusing on relevant features for necking predictions in hot strip mills.
    \item \textbf{TabNet \cite{arik2021tabnet}:} The TabNet architecture is a type of deep learning model designed specifically for tabular data. It employs a combination of sequential attention mechanisms and feature transformations to extract meaningful patterns from tabular datasets. By iteratively selecting and attending to informative features, TabNet can effectively learn complex relationships within the data, making it well-suited for tasks such as classification and regression on tabular data.}
    \item \textbf{XGBoost \cite{XGBoost}}: XGBoost is an ensemble learning algorithm based on decision trees, particularly gradient boosting. It sequentially builds a series of decision trees, with each subsequent tree attempting to correct the errors of the previous ones. By combining the predictions of multiple weak learners, XGBoost creates a strong predictive model that can handle both regression and classification tasks effectively. Its ability to handle missing values, regularization techniques, and flexibility in objective functions make it a popular choice for structured/tabular data analysis, achieving state-of-the-art results in various real-world applications.
    \s{\item \textbf{TabPFN \cite{hollmann2025accurate}}: TabPFN is a Transformer-based approach trained via meta-learning to quickly solve small tabular classification problems. Instead of iteratively fitting a model per dataset, TabPFN uses a single forward pass to produce predictions. This is achieved by training on a large distribution of synthetic tasks, enabling near-instant inference for new, small-scale tabular problems.}
\end{itemize}

\begin{table*}[]
\setlength{\tabcolsep}{1.5pt}
\centering
\caption{Performance Comparison Across Models and Tasks \s{Using MAE\%, PAM-5, PAM-2.5, and EV Metrics. The last column ($\Delta_M$) shows the multitask learning gain over the STL baseline.}}
\label{tab:comparison}
\resizebox{1.0\textwidth}{!}{%
\begin{tabular}{cccccccccccccccccccccccccccc}
\toprule
\multirow{2}{*}{Models} &  & \multicolumn{4}{c}{Lower Yield Strength} &  & \multicolumn{4}{c}{0.2\% OS Yield Strength} & \multirow{2}{*}{} & \multicolumn{4}{c}{0.5\% Ext Yield Strength} & \multicolumn{1}{c}{} & \multicolumn{4}{c}{Ultimate Tensile Strength} & \multicolumn{1}{c}{} & \multicolumn{4}{c}{2" Elongation} & \multicolumn{1}{c}{} & \multicolumn{1}{c}{\multirow{2}{*}{\begin{tabular}[c]{@{}c@{}}$\Delta_M$ \\ (↑)\end{tabular}}} \\
\cline{3-6}\cline{8-11}\cline{13-16}\cline{18-21}\cline{23-26}\\[-2ex]
 &  & \begin{tabular}[c]{@{}c@{}}MAE\% \\ (↓)\end{tabular} & \begin{tabular}[c]{@{}c@{}}PAM 5 \\ (↑)\end{tabular} & \begin{tabular}[c]{@{}c@{}}PAM 2.5\\ (↑)\end{tabular} & \begin{tabular}[c]{@{}c@{}}EV\\ (↑)\end{tabular} &  & \begin{tabular}[c]{@{}c@{}}MAE\% \\ (↓)\end{tabular} & \begin{tabular}[c]{@{}c@{}}PAM 5\\ (↑)\end{tabular} & \begin{tabular}[c]{@{}c@{}}PAM 2.5 \\ (↑)\end{tabular} & \begin{tabular}[c]{@{}c@{}}EV\\ (↑)\end{tabular} &  & \multicolumn{1}{c}{\begin{tabular}[c]{@{}c@{}}MAE\% \\ (↓)\end{tabular}} & \multicolumn{1}{c}{\begin{tabular}[c]{@{}c@{}}PAM 5 \\ (↑)\end{tabular}} & \multicolumn{1}{c}{\begin{tabular}[c]{@{}c@{}}PAM 2.5 \\ (↑)\end{tabular}} & \multicolumn{1}{c}{\begin{tabular}[c]{@{}c@{}}EV \\ (↑)\end{tabular}} &  & \multicolumn{1}{c}{\begin{tabular}[c]{@{}c@{}}MAE\% \\ (↓)\end{tabular}} & \multicolumn{1}{c}{\begin{tabular}[c]{@{}c@{}}PAM 5 \\ (↑)\end{tabular}} & \multicolumn{1}{c}{\begin{tabular}[c]{@{}c@{}}PAM 2.5 \\ (↑)\end{tabular}} & \multicolumn{1}{c}{\begin{tabular}[c]{@{}c@{}}EV \\ (↑)\end{tabular}} &  & \multicolumn{1}{c}{\begin{tabular}[c]{@{}c@{}}MAE\% \\ (↓)\end{tabular}} & \multicolumn{1}{c}{\begin{tabular}[c]{@{}c@{}}PAM 5 \\ (↑)\end{tabular}} & \multicolumn{1}{c}{\begin{tabular}[c]{@{}c@{}}PAM 2.5 \\ (↑)\end{tabular}} & \multicolumn{1}{c}{\begin{tabular}[c]{@{}c@{}}EV \\ (↑)\end{tabular}} &  & \multicolumn{1}{c}{} \\
\cline{1-1}\cline{3-6}\cline{8-11}\cline{13-16}\cline{18-21}\cline{23-26}\cline{28-28}\\[-2ex]
STL &  & 2.963 & 83.42 &  52.69 & 96.15 &  & 2.858 & 84.96 & 54.08 &  96.41 &  & 3.180 & 84.58 & 51.58 & 94.85 &  & 2.012 & 94.81 & 70.16 & 96.84 &  & 3.762 & 73.85 & 43.30 & 90.00 &  & +0.00 \\
MTL &  & 3.026 & 83.09 & 50.16 & 96.11 &  & 2.921 & 82.25 & 52.27 & 96.40 &  & 3.187 & 83.95 & 50.74 & 94.89 &  & 2.099 & 93.79 & 67.88 & 97.05 &  & 3.987 & 69.81 & 41.29 & 88.89 &  & -2.91\\
MMoE & & 2.933 & 84.11 & 53.60 & 96.03 &  & 2.836 & 85.29 & 54.53 & 96.36 &  & 3.148 & 84.08 & 52.81 & 94.89 &  & 2.018 & 94.32 & 70.46 & 97.15 &  & 3.790 & 72.43 & 43.36 & 89.81 &  & +0.35 \\
PLE &  & 2.924 & 84.26 & 53.82 & 96.06 &  & 2.819 & 85.03 & 55.33 & 96.43 &  & 3.164 & 83.52 & 52.96 & 94.57 &  & 1.980 & 94.59 & 71.20 & 97.15 &  & 3.826 & 72.25 &  43.03 & 89.57 &  & +0.61 \\
STEM &  & 2.927 & 84.16 & 52.58 & 96.27 &  & 2.848 & 84.89 & 54.07 & 96.21 &  & 3.126 & 83.87 & 52.19  & 94.67 &  & 2.014 & 94.19 & 69.95 & 97.15 &  & 3.956 & 70.77 &  41.47 & 88.80 &  & -0.40 \\
FT-Transformer &  & 2.990 & 82.81 & 52.42 & 96.20 &  & 2.890 & 84.23 & 53.24 & 96.44 &  & 3.220 & 82.51 & 51.49 & 94.82 &  & 2.159 & 92.91 & 66.39 & 97.02 &  & 3.864 & 71.88 & 41.93 & 89.74 &  & -2.66 \\
SAINT &  & 2.964 & 83.15 & 52.45 & 96.25 &  & 2.915 & 84.08 & 53.43 & 96.23 &  & 3.207 & 82.75 & 51.72 & 94.81 &  & 2.021 & 94.57 & 70.03 & 97.18 &  & 3.781 & 72.86 & 43.70 & 90.03 &  & -0.77 \\
MultiTab &  & 2.922 & 83.89 & 53.74 & 96.26 &  & 2.849 & 84.66 & 54.47 & 96.52 &  & 3.103 & 84.17 & 53.00 & 95.01 &  & 1.992 & 94.86 & 70.85 & 97.25 &  & 3.915 & 71.25 & 42.17 & 89.26 &  & +0.21 \\
XGBoost &  & 2.732 & 85.96 & 57.89 & 96.53 & & 2.668 & 86.66 & 58.56 & 96.75 & & 2.976 & 85.61 & 56.55 & 94.96 & & 1.780 & 95.59 & 76.42 & 97.64 & & 3.724 & 73.86 & 44.34 & 90.29 & & +6.68 \\
\midrule
TabPFN (n.f.t) & & 2.697 & 86.52 & 57.56 & 96.72 & & 2.654 & 87.03 & 57.69 & 96.93 &  & 2.917 & 86.42 & 55.60 & 95.41 &  & 1.751 & 95.94 & 76.75 & 97.75 &  & 3.656 & 74.52 & 46.00 & 90.79 &  & +8.04\\
TabPFN (s.f.t) & & 2.652 & \textbf{86.94} & 58.46 & \underline{96.77} & & 2.614 & 87.54 & 58.23 & 96.97 &  & 2.901 & 86.75 & 56.24 & 95.43 &  & \textbf{1.701} & \textbf{96.26} & \textbf{78.14} & \textbf{97.83} &  & 3.642 & 74.64 & 46.01 & 90.83 &  & +9.29\\
\midrule
TabPFN (m.f.t.) & & \underline{2.647} & \underline{86.93} & \underline{58.56} & 96.76 & & \underline{2.604} & \textbf{87.75} & \underline{58.66} & \underline{96.98} &  & \underline{2.874} & \underline{87.13} & \underline{56.92} & \textbf{95.46} &  & \underline{1.704} & 96.17 & 77.79 & \underline{97.82} &  & \underline{3.623} & \underline{74.73} & \textbf{46.19} & \underline{90.93} &  & \underline{+9.64}\\
TabPFN (m.a.f.t.) & & \textbf{2.646} & \underline{86.93} & \textbf{58.77} & \textbf{96.78} & & \textbf{2.603} & \underline{87.66} & \textbf{58.78} & \textbf{96.99} &  & \textbf{2.870} & \textbf{87.19} & \textbf{57.04} & \textbf{95.46} &  & 1.705 & \underline{96.23} & \underline{77.92} & \underline{97.82} &  & \textbf{3.611} & \textbf{75.56} & \underline{46.12} & \textbf{90.98} &  & \textbf{+9.73}\\
\bottomrule
\end{tabular}}
\end{table*}

\subsection{Implementation Details}
All experiments were conducted using PyTorch Lightning\s{~\cite{Falcon_PyTorch_Lightning_2019}} and trained on a single NVIDIA RTX A5500 GPU. Hyperparameter tuning for all baselines were performed using the Bayesian Hyperband\s{~\cite{falkner2017combining}} search algorithm via Weights \& Biases (WandB). Training hyperparameters were swept over optimizers \{Adam\s{~\cite{kingma2014adam}}, SGD\s{~\cite{lecun2002efficient}}\}, learning rates \{1e-4, 1e-3, 1e-2, 1e-1\}, dropout rates \{0.0, 0.1, 0.2, 0.3\}, weight decay values \{0, 1e-5, 1e-4, 1e-3\}, and batch sizes \{32, 64\}. All models were trained for 50 epochs using a Cosine Annealing learning rate scheduler\s{\cite{loshchilov2016sgdr}} for smooth convergence. Model-specific hyperparameters, such as the number of layers \{1, 2, 3\}, feed-forward dimensions \{64, 128, 256, 512\}, feature embedding sizes \{8, 16, 32\}, and the number of experts for mixture-of-experts models \{1, 2, 4, 8\}, were also tuned. For all experiments involving finetuning TabPFN\cite{hollmann2025accurate}, we used a time limit of 120 seconds, a learning rate of 1e-5, a batch size of 8, and disabled preprocessing at inference time to match fine-tuning. For multitask adapter fine-tuning, we used $s=15$. Each model was trained using five random seeds to ensure robustness, and all performance metrics were averaged across these runs. All baselines were either implemented using their official codebases or closely reproduced based on original descriptions.

\section{Results and Discussion}
Table \ref{tab:comparison} presents the performance of various baseline models alongside our proposed TabPFN variants on five target properties, explained in Section \ref{Tasks Overview}. For each property, we report four evaluation metrics: \textbf{Mean Absolute Error (MAE\%)}, \textbf{Predictive Accuracy Metric (PAM)} at 5\% and 2.5\% thresholds (PAM-5 and PAM-2.5), and \textbf{Explained Variance (EV)}. The final column shows the overall multitask learning gain \(\Delta_M\), calculated based on MAE.
Among the baselines, several models exhibit performance degradation compared to STL (Single Task Learning), as reflected by their negative $\Delta_M$ values. In fact, MTL yields the lowest overall $\Delta_M$ at –2.91, indicating that naive multitasking without effective task coordination may hinder performance. Models such as Mixture-of-Experts (MMoE) and PLE show more consistent improvements across tasks, achieving positive $\Delta_M$ values and better balance in MAE and PAM metrics. In contrast, STEM, while competitive in some individual metrics, fails to achieve an overall gain, reflected by its negative $\Delta_M$. Transformer-based baselines like FT-Transformer and SAINT underperform in this small-data regime, particularly on tasks like Elongation and 0.5\% Ext Yield Strength. This aligns with existing findings that transformer models trained from scratch on tabular data often struggle to match the performance of simpler MLP-based models when limited data is available.

To evaluate TabPFN, we used four different settings. The first setting, \textbf{TabPFN (n.f.t.)}, refers to using the off-the-shelf model with \textbf{n}o \textbf{f}ine-\textbf{t}uning on downstream tasks. Despite the absence of task-specific fine-tuning, the standard TabPFN outperforms all other baselines across all tasks, highlighting the impressive capabilities of this foundational model.

The second setting, \textbf{TabPFN (s.f.t.)}, applies \textbf{s}ingle-task \textbf{f}ine-\textbf{t}uning on the target dataset. It improves upon the non-fine-tuned version across most MAE scores and achieves an impressive 
 \(\Delta_m = +9.29\), confirming the benefit of task-specific adaptation. However, this approach requires fine-tuning and storing a different model for each of our five tasks. Each fine-tuned model also does not have any notion of cross-task awareness, which is a potential inhibitor of further performance improvements. We aim to address this issue with our proposed multitask fine-tuning settings.

The third setting, \textbf{TabPFN (m.f.t.)}, uses our proposed \textbf{m}ulti-task \textbf{f}ine-\textbf{t}uning strategy that uses the average of all the task regression targets as the training signal during fine-tuning. Fine-tuning with this signal embedded with multitask information outperforms all baselines and TabPFN (s.f.t) across nearly all metrics and tasks. As a result, the multitask gain rises to $\Delta_M = +9.64$, reflecting a clear benefit from using a multitask aware training signal. It is quite fascinating that TabPFN achieves better task-specific performance despite not being fine-tuned on the corresponding task-specific signal. In addition to achieving great overall performance, this setting is five times more efficient regarding training time and memory storage since we only fine-tune and store a single model.

The fourth and final setting, \textbf{TabPFN (m.a.f.t.)}, uses \textbf{m}ultitask \textbf{a}dapter \textbf{f}ine-\textbf{t}uning which integrates adapter layers to provide task-specific supervised auxiliary training signals during fine-tuning. These auxiliary signals are introduced in addition to the main regression target used in m.f.t (i.e., the average of the task targets). This setting performs similarly to m.f.t. but slightly improves performance in several PAM-2.5 and EV scores. Overall, it achieves the best multitask gain among all variants of TabPFN with $\Delta_M=+9.73$. The majority of the improvement seems to be from the Elongation task, which is the least correlated among the other tasks. This suggests that standard m.f.t is great for strongly correlated tasks, but adding the task-specific supervised signals strongly benefits less correlated targets. Similar to m.f.t, this setting also offers substantial efficiency benefits by requiring the fine-tuning and storage of a single TabPFN model, further validating the effectiveness of the adapter-based MTL strategy. 

\begin{figure}
    \centering
    \includegraphics[width=\linewidth]{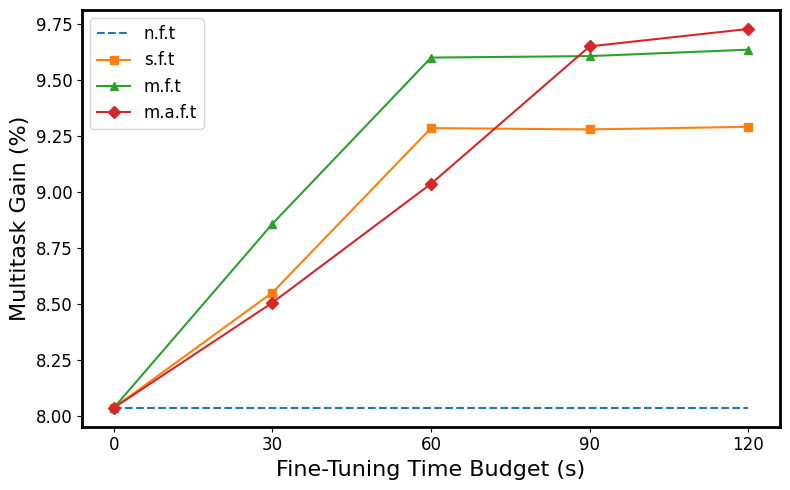}
    \caption{Multitask gain ($\Delta_m$) as a function of fine-tuning time budget for four strategies: no fine-tuning (n.f.t), single-task fine-tuning (s.f.t), multitask fine-tuning (m.f.t), and multitask adapter fine-tuning (m.a.f.t).}
    \label{fig:strategies}
\end{figure}

Figure \ref{fig:strategies} provides further details about the evolution of multitask gain as a function of the fine-tuning time budget. As we can see, all fine-tuning methods significantly improve multitask gain, with m.f.t always leading s.f.t. In our setting, m.f.t and s.f.t performance plateaus after 60 seconds, while m.a.f.t continues to improve, achieving the best multitask gain through 120 seconds of fine-tuning.

In summary, the TabPFN-based models—especially the proposed multi-task fine-tuning variants—consistently outperform both classical and modern baselines across all tasks and metrics. The use of multi-task fine-tuning with the average task signal alongside task-specific auxiliary signals yields the best performance for a fraction of the computational resources than single-task fine-tuning, supporting the practicality of our approach for real-world tabular regression applications.

\section{Conclusion}
This paper introduced a novel multitask fine-tuning framework for TabPFN, a foundation model originally developed for single-task classification and regression on tabular data, and demonstrated its effectiveness for the prediction of multiple mechanical properties in the Thin Slab Direct Rolling (TSDR) steel manufacturing process. By extending TabPFN through two multitask learning strategies—task-averaged fine-tuning and adapter-based fine-tuning—we enabled the model to generalize across multiple regression targets without compromising its lightweight, inference-time efficiency.

Our experiments on an industrial dataset\s{ from Algoma Steel Inc.} revealed that both multitask variants consistently outperform classical machine learning baselines and modern deep multitask architectures across a range of metrics. Remarkably, the averaging-based fine-tuning approach retains the original structure of TabPFN, requiring no additional complexity during inference, while the adapter-enhanced version achieves the highest overall performance with minimal computational overhead.

These results underscore the potential of adapting foundation models like TabPFN for high-stakes industrial applications where data efficiency, scalability, and predictive accuracy are critical. The proposed approach not only reduces the training and storage burden compared to traditional single-task fine-tuning pipelines but also enhances model robustness and generalization across diverse tasks. Ultimately, such models can serve as reliable tools for real-time quality control and process optimization within modern steel manufacturing environments.
\s{\section*{Acknowledgment}}

\bibliographystyle{IEEEtran}
\bibliography{references}

\s{\begin{IEEEbiography}[{\includegraphics[width=1in,height=1.25in,clip,keepaspectratio]{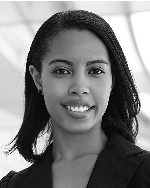}}]{First A. Author} (M'76--SM'81--F'87) and all authors may include 
biographies. Biographies are often not included in conference-related
papers. This author became a Member (M) of IEEE in 1976, a Senior
Member (SM) in 1981, and a Fellow (F) in 1987. The first paragraph may
contain a place and/or date of birth (list place, then date). Next,
the author's educational background is listed. The degrees should be
listed with type of degree in what field, which institution, city,
state, and country, and year the degree was earned. The author's major
field of study should be lower-cased. 

The second paragraph uses the pronoun of the person (he or she) and not the 
author's last name. It lists military and work experience, including summer 
and fellowship jobs. Job titles are capitalized. The current job must have a 
location; previous positions may be listed 
without one. Information concerning previous publications may be included. 
Try not to list more than three books or published articles. The format for 
listing publishers of a book within the biography is: title of book 
(publisher name, year) similar to a reference. Current and previous research 
interests end the paragraph. The third paragraph begins with the author's 
title and last name (e.g., Dr.\ Smith, Prof.\ Jones, Mr.\ Kajor, Ms.\ Hunter). 
List any memberships in professional societies other than the IEEE. Finally, 
list any awards and work for IEEE committees and publications. If a 
photograph is provided, it should be of good quality, and 
professional-looking. Following are two examples of an author's biography.
\end{IEEEbiography}

\begin{IEEEbiography}[{\includegraphics[width=1in,height=1.25in,clip,keepaspectratio]{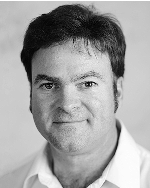}}]{Second B. Author} was born in Greenwich Village, New York, NY, USA in 
1977. He received the B.S. and M.S. degrees in aerospace engineering from 
the University of Virginia, Charlottesville, in 2001 and the Ph.D. degree in 
mechanical engineering from Drexel University, Philadelphia, PA, in 2008.

From 2001 to 2004, he was a Research Assistant with the Princeton Plasma 
Physics Laboratory. Since 2009, he has been an Assistant Professor with the 
Mechanical Engineering Department, Texas A{\&}M University, College Station. 
He is the author of three books, more than 150 articles, and more than 70 
inventions. His research interests include high-pressure and high-density 
nonthermal plasma discharge processes and applications, microscale plasma 
discharges, discharges in liquids, spectroscopic diagnostics, plasma 
propulsion, and innovation plasma applications. He is an Associate Editor of 
the journal \emph{Earth, Moon, Planets}, and holds two patents. 

Dr. Author was a recipient of the International Association of Geomagnetism 
and Aeronomy Young Scientist Award for Excellence in 2008, and the IEEE 
Electromagnetic Compatibility Society Best Symposium Paper Award in 2011. 
\end{IEEEbiography}

\begin{IEEEbiography}[{\includegraphics[width=1in,height=1.25in,clip,keepaspectratio]{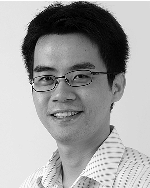}}]{Third C. Author, Jr.} (M'87) received the B.S. degree in mechanical 
engineering from National Chung Cheng University, Chiayi, Taiwan, in 2004 
and the M.S. degree in mechanical engineering from National Tsing Hua 
University, Hsinchu, Taiwan, in 2006. He is currently pursuing the Ph.D. 
degree in mechanical engineering at Texas A{\&}M University, College 
Station, TX, USA.

From 2008 to 2009, he was a Research Assistant with the Institute of 
Physics, Academia Sinica, Tapei, Taiwan. His research interest includes the 
development of surface processing and biological/medical treatment 
techniques using nonthermal atmospheric pressure plasmas, fundamental study 
of plasma sources, and fabrication of micro- or nanostructured surfaces. 

Mr. Author's awards and honors include the Frew Fellowship (Australian 
Academy of Science), the I. I. Rabi Prize (APS), the European Frequency and 
Time Forum Award, the Carl Zeiss Research Award, the William F. Meggers 
Award and the Adolph Lomb Medal (OSA).
\end{IEEEbiography}}

\end{document}